%% file: main.tex
\documentclass[letterpaper]{article}
\usepackage{aaai}
\usepackage{times}
\usepackage{helvet}
\usepackage{courier}
\usepackage{graphicx}
\usepackage[USenglish]{babel}
\usepackage[super]{nth}
\usepackage{float}
\usepackage{subcaption}
\usepackage{pdfpages,multirow,ragged2e}
\usepackage{booktabs}
\usepackage{threeparttable}
\usepackage{flushend}
\usepackage{mathrsfs}   
\usepackage{amsmath}
\usepackage{amssymb}
\usepackage{cleveref}
\usepackage{longtable}
\usepackage[inline]{enumitem}

\newcommand{\regtm}{\texttrademark}

\begin{document}
%
\title{AI-Enabled Operations at Fermi Complex: Multivariate Time Series Prediction for Outage Prediction and Diagnosis}
\author{Milan Jain, Burcu O. Mutlu, Caleb Stam, Jan Strube\textsuperscript{1}\\
Pacific Northwest National Laboratory, Richland, WA, USA\\
\{milan.jain, burcu.mutlu, caleb.stam, jan.strube\}@pnnl.gov\\
\textsuperscript{1}also at University of Oregon, Eugene, OR, USA
\AND
Brian A. Schupbach, Jason M. St. John, William A. Pellico\\
Fermi National Accelerator Laboratory, Batavia, IL, USA\\
\{schupbach, stjohn, pellico\}@fnal.gov
}
\maketitle
\begin{abstract}
\begin{quote}

The Main Control Room of the Fermilab accelerator complex continuously gathers extensive time-series data from thousands of sensors monitoring the beam. However, unplanned events such as trips or voltage fluctuations often result in beam outages, causing operational downtime. This downtime not only consumes operator effort in diagnosing and addressing the issue but also leads to unnecessary energy consumption by idle machines awaiting beam restoration. The current threshold-based alarm system is reactive and faces challenges including frequent false alarms and inconsistent outage-cause labeling. To address these limitations, we propose an AI-enabled framework that leverages predictive analytics and automated labeling. Using data from $2,703$ Linac devices and $80$ operator-labeled outages, we evaluate state-of-the-art deep learning architectures, including recurrent, attention-based, and linear models, for beam outage prediction. Additionally, we assess a Random Forest-based labeling system for providing consistent, confidence-scored outage annotations. Our findings highlight the strengths and weaknesses of these architectures for beam outage prediction and identify critical gaps that must be addressed to fully harness AI for transitioning downtime handling from reactive to predictive, ultimately reducing downtime and improving decision-making in accelerator management.
\end{quote}
\end{abstract}

\section{Introduction}

The Fermilab accelerator complex is the United States' flagship facility for High Energy Physics (HEP). The complex consists of a $400$~MeV proton Linac, an $8$~GeV rapid-cycling synchrotron Booster, a $150$~GeV ramped Main Injector, an $8$~GeV Recycler storage ring, the Muon Campus Delivery Ring, two high-power Neutrino target systems, $120$~GeV fixed target beam lines and many associated transfer lines. The operation and maintenance of these dozen or so ``machines" and their associated systems requires approximately 400 Accelerator Division employees plus outside contractors and auxiliary laboratory support personnel and activities. The accelerator systems vary in age, constituent technology, and mode of operation, and their management is made possible by a sophisticated controls infrastructure. While legacy software tools have been successful in meeting laboratory needs over the lab's fifty years, modern computing resources and advances in Artificial Intelligence (AI) techniques applied to large data sets are required to transform the management of such large complexes for the better. Subsequently, many organizations operating similarly large facilities have moved from a reactive approach to a ``data first" approach, using copiously available data to improve real-world practices. 

In an accelerator complex, the control room monitors the beam and responds to situations when the beam goes down. For instance, for the linear accelerator (Linac) alone, the control system monitors and issues commands to $> 4000$ control system parameters at frequencies ranging from $15~Hz$ to once every few minutes. Two different permit signals control the presence of beam in the upstream and downstream sections of the Linac, respectively. The beam could be absent for several known and unknown reasons, the latter of which operators have to spend time to investigate. When either beam permit goes down, the operators intervene to first ensure that it is not a false alarm.
Once operators confirm the outage, they gather information by visually analyzing data and determine the right action to rectify the issue. A beam outage leads to downtime causing wasted time and energy~\cite{jain2022cape,strubeartificial}.


\begin{figure*}
    \centering
    \includegraphics[trim={0 3.2cm 0 3.2cm},clip, width=0.9\linewidth]{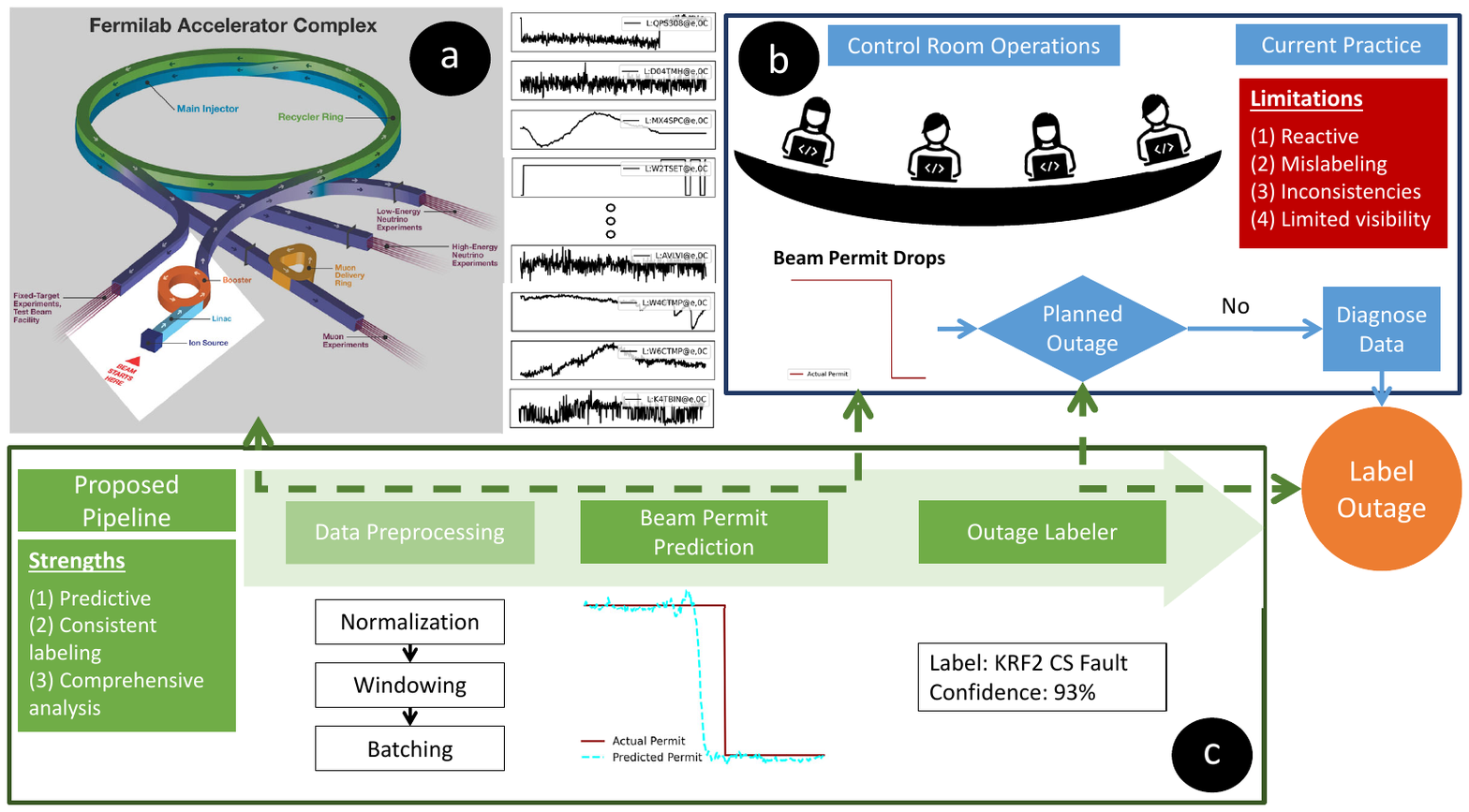}
    \caption{Overview diagram illustrating: (a) the Fermilab accelerator complex with sample device data collected from the Linac, (b) current operations in the FNAL control room, highlighting key limitations, and (c) the proposed predictive maintenance pipeline along with its potential benefits.}
    \label{fig:complex}
\end{figure*}

However, this legacy way of monitoring the beam and responding to an outage poses four major concerns: (1) false alarms waste operator time, (2) it is reactive rather than predictive, (3) the number of devices to monitor and the amount of data exceed human capacity to process, and (4) inconsistent or incorrect labeling complicates bookkeeping and higher-order analytics of faults.

Even during normal beam operations, the operations staff are notified of approximately $15,000$ alarms daily, along with additional status indicators, a number that can increase significantly during study periods and certain operational phases. Usually, this \emph{high rate of false alarms} distracts operators from concerns (also defined as precursors) that could indicate an impending outage. And since the response is mostly \emph{reactive}, operators usually have little time to avert the outage. 

Once a beam outage begins, certain measures can be taken so that, while staff work to restore operation, those machines awaiting the beam may have their power consumption reduced. This could be a partial or complete operating power reduction, depending upon context. Such power conservation is normal practice in simple cases, but requires the judgment of a human expert as to the likely duration of the outage, and the benefit to be gained by taking steps which must then be reversed when the outage ends. However, this judgment relies on the examination of those devices \emph{potentially} related to the outage one by one. The list of devices to analyze is often derived from intuition and past experience. The time spent by human operators on pattern-hunting from \emph{large-scale data} to validate and understand anomalous conditions currently occupies a significant portion of their response time to a beam outage. And that is where proper labeling of the outage could be helpful as it can provide a rough estimate of outage duration. Currently, the outage is labeled subjectively and can result in \emph{inconsistent or incorrect labeling}.

We exploit the predictive power of AI applied to copiously available time-series data generated by the Fermilab accelerator control system to go beyond the present system of threshold-based alarms. With appropriately selected and prepared data, algorithmically optimized, nonlinear functions learned on our large datasets can detect and predict emerging anomalous conditions not visible to traditional alarm-setting methods. The proposed pipeline (see bottom half of \Cref{fig:complex}) enables the use of ML/AI that augments the data flow to the control room with analytics of outages, reduces the time to label them meaningfully and minimizes the number of incorrect or inconsistent labels.

The task of predictive monitoring of beam is formulated as predicting the status of upstream and downstream Linac permit bits using time series data from thousands of Linac devices. The predictive power of deep learning (DL) methods can level up the operations by finding precursors related to an anomaly for predicting an outage and help operators transition from being reactive to \emph{being predictive}. However, to avoid \emph{alarm fatigue}, we must reduce the number of false positives while maintaining good predictive power. In this study, we evaluate state-of-the-art (SOTA) DL architectures---spanning recurrent, attention-based, and linear models---for beam-outage prediction. 

For the second task of automatically labeling outage causes, we trained and evaluated a Random Forest model that labels an outage when either beam permit goes down, along with a label confidence score. 

The key contributions of this work are summarized below:
\begin{enumerate}
    \item A thorough data collection campaign from the FNAL accelerator complex, including manual labeling of outages by operators, ensuring high-quality datasets for analysis.
    \item The comprehensive evaluation of SOTA multivariate time-series prediction DL architectures employed for beam-permit prediction. 
    \item A performance evaluation of an automated outage labeler.
    \item The detailed discussion of key findings and lessons learned to inform and guide future efforts in this domain. 
\end{enumerate}

\section{Related Work}
The literature on deep learning for operational efficiency and reliability in large-scale facilities can be broadly classified into two categories based on the timing of the response: (1) anomaly detection and diagnosis, and (2) predictive maintenance. This section reviews SOTA techniques proposed in the literature for both categories. 

\subsubsection{Anomaly Detection and Diagnosis}
Anomaly detection and diagnosis (AD\&D) is a reactive method that involves near real-time analysis of massive data streams from thousands, or even millions, of sensors to identify unexpected behavior with precision and timeliness. A detailed review of time series anomaly detection can be found in \cite{chandola2009anomaly,schmidl2022anomaly}. 
Recent advancements include models such as LSTM-VAE~\cite{park2018multimodal}, MSCRED~\cite{zhang2019deep}, and TAnoGAN~\cite{bashar2020tanogan}. These architectures predominantly focus on reconstructing input sequences, training models exclusively on normal data with the assumption that anomalous instances will be poorly reconstructed and thus stand out. 

Detecting anomalies in complex time series data is often challenging due to their context-dependent nature, as anomalies can encompass any unusual, irregular, or unexpected observations, making it difficult to establish a clear definition. Yang et al.~\cite{yang2023dcdetector} proposed DCdetector, a dual-attention representation-learning architecture that uses contrastive learning to produce representations capable of effectively distinguishing individual instances, offering a promising approach for time series anomaly detection. Another recent study proposed TFAD~\cite{zhang2022tfad} -- a time-frequency domain analysis time series anomaly detection model utilizing both time and frequency domains for performance improvement.

Despite these advancements, a key limitation of AD\&D techniques is their reactive nature. While they excel at diagnosing issues, they fall short in facilitating proactive measures to prevent outages or address potential problems before they arise, a critical requirement for this study.

\subsubsection{Predictive Maintenance}
Predictive maintenance, in contrast, forecasts the system's future state over single or multiple time steps based on data in a look-back window, enabling proactive fault diagnosis through accurate predictions. 
Traditionally, statistical models such as ARIMA, AR, and exponential smoothing models~\cite{box2015time} have been used for univariate time series forecasting. However, their inability to capture multivariate nonlinear relationships between variables, and the increasing availability of ``big" data and compute, led to the rise of deep learning methods, such as recurrent networks (e.g. LSTM~\cite{hochreiter1997long}) for multivariate time-series forecasting. Further, these DL-based forecasting architectures were extended to handle auxiliary information, such as static and time-varying features. 

Inspired by the success of attention mechanisms in natural language processing (NLP) and computer vision (CV)~\cite{vaswani2017attention}, researchers began integrating attention into time series forecasting to capture long-term dependencies. For example, a dual-stage attention-based recurrent neural network was introduced for time series prediction~\cite{qin2017dual}, while the Temporal Fusion Transformer~\cite{lim2021temporal} was proposed to merge high-performance multi-horizon forecasting with interpretable temporal dynamics. Nevertheless, transformer models face challenges with high computational costs, prompting the development of more efficient variants. Informer~\cite{zhou2021informer} and Autoformer~\cite{wu2021autoformer} address efficiency bottlenecks through memory-efficient attention designs for long-term forecasting, while FEDformer~\cite{zhou2022fedformer} and FiLM~\cite{zhou2022film} employ Fast Fourier Transformation to enhance the extraction of long-term dependencies. Despite the advancements, recent studies~\cite{zeng2023transformers} reveal that linear models, such as N-BEATS~\cite{oreshkinn}
and TS-Mixer~\cite{ekambaram2023tsmixer}, 
can outperform attention-based networks for multivariate forecasting. 

Although predictive maintenance has been extensively studied in areas such as energy efficiency~\cite{jain2019beyond}, manufacturing~\cite{ccinar2020machine}, and several other industrial systems~\cite{nguyen2019new,shiva2024anomaly}, there has been little to no comprehensive evaluation of state-of-the-art multivariate time series models for beam outage prediction and labeling using real-world data.

\section{Problem Formulation}
\subsubsection{Beam-Permit Prediction}
The control system data for the Fermilab accelerator complex comes in the form of a time series with $N$ analogue devices, $D$ digital bit devices, and $F$ future covariates. The presence of beam is controlled by a \emph{permit}, with 1 allowing beam and 0 prohibiting beam, that combines threshold-based information from this data stream to ensure safe operations.
Our task is to learn $\mathscr{F}$ (Eq~\ref{eq:bit_prediction}) that can predict the beam permit, $\hat{X}_{bp} \in \mathbb{R}^{L_f \times 1}$, $bp \in D$ for the look-forward window $L_f$, given the historical observations from analogue devices, $\mathbf{X_n} \in \mathbb{R}^{L_b \times N}$, and historical and future data from future covariates, $\mathbf{X}_f \in \mathbb{R}^{{(L_b+L_f)} \times F}$, wherein, $L_b$ is the size of the look-back window. 
\begin{equation}
    \mathscr{F}: (\{\mathbf{X_n^t}\}_{t=1}^{L_b}, \{\mathbf{X_f^t}\}_{t=1}^{L_b+G+L_f}) \rightarrow \{\hat{X}_{bp}\}_{t=L_b+G}^{L_b+G+L_f}
    \label{eq:bit_prediction}
\end{equation}


In Equation~\ref{eq:bit_prediction}, we introduce a gap $G$ between the look-back and look-forward windows. This allows us to strike a good balance between the uncertainty due to larger look-forward windows and the increasing uncertainty on the prediction of events further in the future. \Cref{app:sensitivity_analysis} studies the effect of this gap quantitatively. 



\subsubsection{Outage Labeling}
\label{sec:decisionTree}
Once the beam permit goes down, we label the outage cause. The operator's labeling relies on subjective experience rather than standardized nomenclature, resulting in inconsistencies. To address this, we train a classifier $\mathscr{F}_l$ that assigns a label $\mathscr{L}$ to an outage at time $t'$, when the beam permit $X_{bp}$ goes down, i.e., $X_{bp}^{t'}=0$ and $X_{bp}^{(t'-1)}=1$. Our classifier is trained on $\mathbf{X_n} \in \mathbb{R}^{L_b \times (N+D)}$, historical observations from all devices ending at time step $t'$. 
In particular, we take the difference between the data at the time of outage and the average of the data from the last $k$ time steps. Let $\mathscr{F}_{a}: \mathbf{\{X_n^t\}}_{t={(t'-L_b)}}^{t'} \rightarrow \mathbf{X_n}$ be our aggregation function. 

\begin{equation} \mathscr{F}_{a} = {X_n^{t'}}-\left(\frac{1}{k}\sum\limits_{t=(t'-L_b)}^{t'-1}{X_n^t}\right)
\end{equation} 

This formulation tested better than alternatives such as mean aggregation across the entire look-back window, inclusion of a look-forward window, or skipping aggregation and classifying on single time steps. Finally, if $\mathscr{F}_{rf}: \mathbf{X_n} \rightarrow \mathscr{L}$ is our random forest classifier, then

\begin{equation}
    \mathscr{F}_l = \mathscr{F}_{rf} \circ \mathscr{F}_a
\end{equation}

\section{Experimental Setup}
This study focused on data from 2703 devices of the 400 MeV proton Linac, collected in 2024. Every year, the accelerator complex runs for a specific period (usually November--July) with major activities happening between March and July; and shuts down between August and November for maintenance. Data during regular operations is stored in rolling buffers of fixed size. A significant amount of effort went into building and deploying a stable data collection and storage pipeline outside of these buffers. 
Next, we discuss the data pipeline, preprocessing steps, and outage data in detail.

\subsection{Data Collection and Processing}
The accelerator control system's Data Logger nodes record data streams into circular buffers. To store this data for a longer period than the lifetime of the circular buffers, this project developed a data acquisition pipeline that writes the data to long-term storage in \texttt{Parquet}~\cite{Vohra2016} format with the lossless \texttt{snappy} compression~\cite{snappy}. Missing values, e.g., from faulty reads are interpolated using a forward fill algorithm for both offline and online processing.  
The $2703$ devices include $1719$ readings, $842$ settings, and $142$ status bits, and are stored in the \texttt{parquet.snappy} file format once per hour.



\begin{figure}
    \centering
    \includegraphics[width=0.98\linewidth]{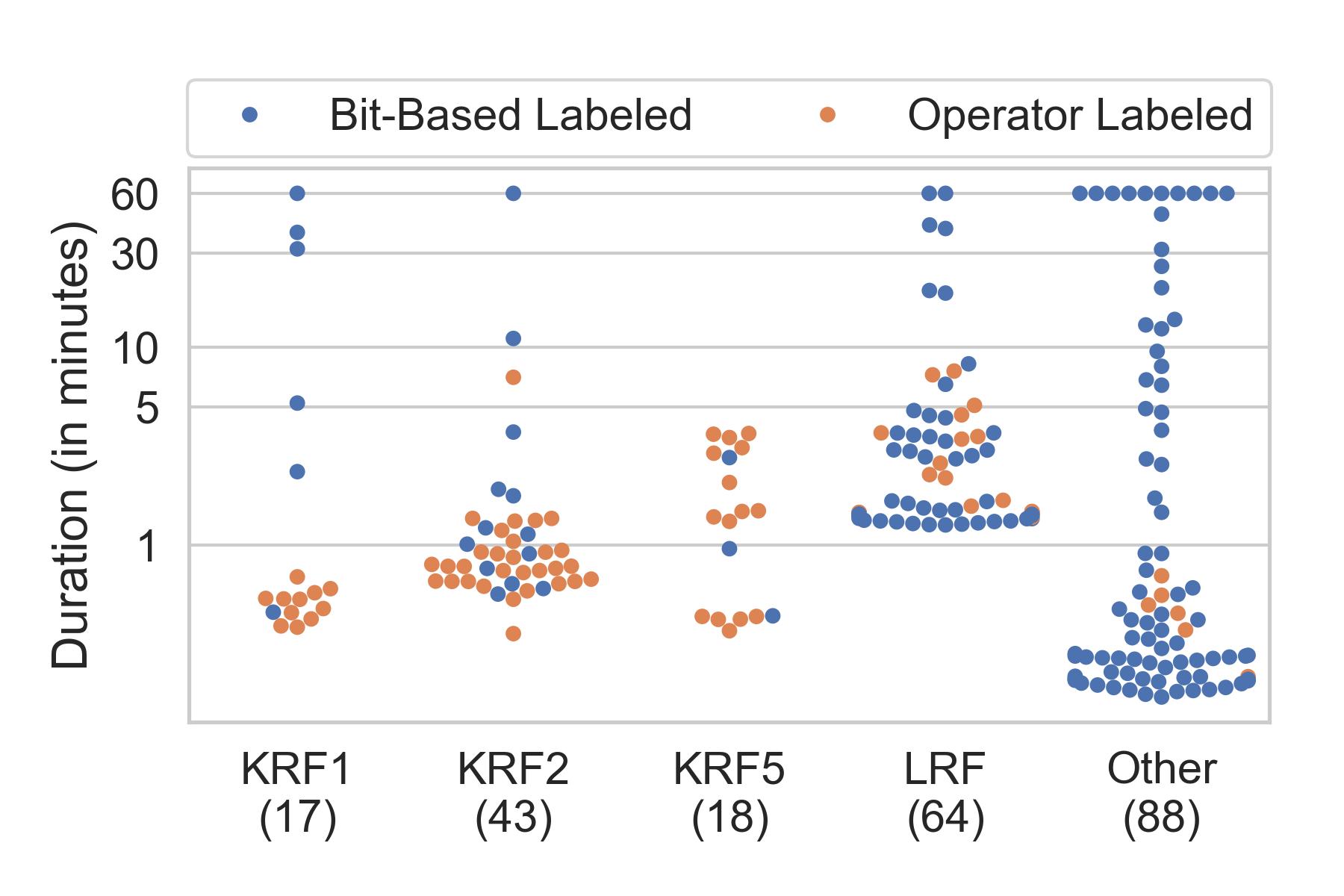}
    \caption{Distribution of outage duration by class. The duration is limited to 60 minutes by the size of a single data file.}
    \label{fig:outage-dist}
\end{figure}

\subsubsection{Outages}
For training our models, we extract windows around the time when the beam permit $X_{bp}$ drops. Because there are frequent fluctuations of the permit, we skip ``outages'' of less than $10$ seconds. We search through the recorded files and save a window of $30$ seconds before the outage starts and $10$ seconds later. We find 205 such occurrences in our data, their causes initially unlabeled, and we explore three different ways of assigning labels.
\begin{description}
    \item[Operator-Labeled Outages] {Operators in the control room investigate the cause of outages and assign labels based on their findings and prior experience, generally only for outages lasting longer than one minute. These are logged with wall clock time and duration in minutes. We match these labels with an outage in our set based on start time and refine the operator-assigned start time and duration with the 15-Hz clock from the outage data. Our data contains 80 outages that were labeled by operators. Labels assigned by operators distinguished 9 outage types occurring in 12 different locations. Labels like “KRF1 CS Fault” give combined information of the location (KRF1) and the outage type (CS Fault), allowing an operator to diagnose and address outages. We used these labels as the basis for our two automated methods described below.}
    \item[Bit-Labeled Outages] {We identified 12 digital devices as providing bit-wise system status. These devices store multiple bits of information, where each bit acts as a binary indicator for a specific system event, including high voltage conditions or spark trip occurrences. We analyzed these devices with respect to the operator-labeled outages and matched patterns of bit flips with operator labels. We observed that within a few seconds of the outage, these devices show distinct patterns for different types of outages. Leveraging this information, we developed a bit-based labeler to categorize outages. This tool enables us to label 74 out of 150 previously unlabeled outages within 2 seconds of an outage (to account for possible time slippage between the permits and the status bits). In addition to helping us with previously unlabeled data, it also enhances our understanding of operator-labeled data by overcoming inconsistencies and ambiguity in human-generated labels.}
    \item[Decision Tree-Labeled Outages]{The digital devices aggregate information from different sources, but they fail to account for higher-order correlations between variables. We have trained on the devices a random-forest classifier that serves as an important cross-check to the bit-based labeler. Additionally, this approach allows us to study specific signatures of a given outage, which we will use to study precursors in the future. The details of the random forest classifier were discussed in Section~\ref{sec:decisionTree}.}
\end{description}

A breakdown of the duration of different outages is shown in Figure~\ref{fig:outage-dist}, demonstrating the power of the automated method that is able to assign labels to many instances without human annotation. We will be further able to refine the ``LRF'' category with additional labeled data. The large number of outages in the ``Other'' category, which also has the largest number of cases without human annotation, suggests that we are missing data that would allow us to characterize these cases.




\subsubsection{Non-Outage Instances}
To avoid a training bias from only training on outages and to reduce the number of false positives, we include in our training data periods without beam outage. This data is created from periods where the beam permit was up for at least $30$ consecutive minutes, cropping a $40$-second window at the $20^{th}$ minute within that window. We do not crop more than one such window from any given one-hour file to avoid overlaps. $427$ such non-outage instances are included in our training data. 

\subsection{Data Loading and Windowing}



\subsubsection{Beam Permit Prediction} 
The data is loaded from the \texttt{parquet.snappy} files and converted into overlapping windows for model training and inference using \texttt{TrainDataset} and \texttt{InferenceDataset} dataloader APIs from \texttt{Darts}~\cite{JMLR:v23:21-1177}. These APIs allow us to specify static, past, and future covariates conveniently.

For the beam-permit prediction model, the look-back window size, $L_b$, is set to $30$ ticks ($2s$), the look-forward window size, $L_f$, is set to $60$ ticks ($4s$), and the gap $G$ between the look-back and the look-forward windows is set to $30$ ticks ($2s$). These values were chosen as the model's performance converged at these settings. Further details on the sensitivity analysis of these parameters can be found in \Cref{app:sensitivity_analysis}. The stride is fixed at $1$, and the feature dimension size is $1719$ because the look-back window only includes analogue devices.

We used $75$ of the $125$ beam-permit-labeled outage instances, $40$ of the $80$ operator-labeled outage instances, and $375$ of the $427$ non-outage instances for the model training. To eliminate training bias, the data files were shuffled prior to training. The validation data consisted of $10$ beam-permit-labeled outage instances and $21$ non-outage instances. The test dataset included the remaining $40$ beam-permit-labeled outage instances, $40$ operator-labeled outage instances, and $31$ non-outage instances. 

\subsubsection{Outage Labeler} For the outage classification task, the look-back window size $L_b$ is set to $6$ ticks ($0.4$s). The feature dimension is $2703$, including both analogue and digital devices for maximal predictive power. These preprocessing settings were selected alongside random forest hyperparameters during tuning. The data for outage classification is limited to the $80$ operator-labeled outage instances. We used $8$-fold cross-validation during hyperparameter tuning. The final model was trained on all $80$ operator-labeled outages for deployment.


\subsection{Implemented Models}
\subsubsection{Bit-Permit Prediction}
We choose SOTA models available in \texttt{Darts} from all three architecture categories: (1) recurrent networks, (2) attention-based networks, and (3) linear networks. In recurrent networks, we trained an LSTM (using \texttt{BlockRNNModel}) with two recurrent layers each with a hidden dimension of $25$. Attention-based networks include the vanilla transformer~\cite{vaswani2017attention} and linear networks include N-BEATS~\cite{oreshkinn}, N-HiTS~\cite{challu2023nhits}, TiDE~\cite{das2023long}, and TSMixer~\cite{ekambaram2023tsmixer}. We adopt default hyperparameters for attention-based and linear-networks.

Though Temporal Fusion Transformer (TFT)~\cite{lim2021temporal}, N-linear, and D-linear~\cite{zeng2023transformers} architectures have demonstrated promising performance on benchmark datasets, we opted not to include them in our study. Despite extensive hyperparameter tuning and implementing recommendations from published studies, we were unable to achieve acceptable results with these architectures. TFT failed to detect any outage, N-linear and D-linear had a false positive rate of 100\%. While open-source libraries offer accessible implementations of advanced deep learning architectures, we believe users would benefit from more clear guidelines on their effective use. 

The subset of outage instances that have operator labels as ground truth is considerably smaller than the set of outage instances that can be used for beam-permit prediction. Therefore, we choose a classical machine learning algorithm, Random Forest~\cite{ho1995random} rather than a neural network-based approach for better robustness in the small data regime. However, while neural architectures such as RNNs and transformers are designed for time series data, random forests expect fixed-size vector inputs. Therefore, our labeler consists of a fixed linear aggregation across the time dimension followed by random forest classification. 

\subsection{Model Training}
\subsubsection{Platform}
The models were trained on a single Nvidia\regtm{} DGX-2 ``Ampere'' A100 GPU (108 SMs) with 40GB HBM2 memory\slash GPU and two-way 128-core AMD EPYC\regtm{} 7742 CPUs at 2.25GHz, 256MB L3 cache, 8 memory channels, and 1TB DDR4 memory. The models were developed using PyTorch 2.4.0+cu12~\cite{NEURIPS2019_9015} and Darts v0.30.0~\cite{JMLR:v23:21-1177} and trained using CUDA 12.0. The Random Forest was developed using scikit-learn~\cite{scikit-learn} and trained on Intel(R) Xeon(R) CPU E5-2620 v4 at 2.10GHz with 16 CPUs and 256K L2 cache.

\subsubsection{Training Parameters}
All the models for beam-permit prediction are trained for 500 epochs with a batch size of 254. For training, we used \texttt{AdamOptimizer} with following parameters: $learning\_rate=0.0005$, $clipnorm=1.0$, and $clipvalue=0.5$. An \texttt{EarlyStopping} callback to monitor \emph{val\_loss} was used with the following parameters: $min\_loss=1e-06$, $patience=10$, $mode=min$, and $restore\_best\_weights=True$. We also employed \texttt{torch.optim.lr\_scheduler.ExponentialLR} with $gamma=0.999$ to adjust the learning rate as training converged. For the random forest, we use the classifier from \texttt{sklearn.ensemble.RandomForestClassifier} with $n\_estimators=200$ and $min\_samples\_split=2$.

\subsubsection{Loss Function}
\emph{Mean squared error} (MSE) between the actual beam permit and the predicted beam permit is the loss function for training beam-permit prediction model. Random forest uses the \emph{Gini impurity} as the split criterion.

%


\input{pred_acc}
\section{Results}
The performance of various architectures was assessed based on two primary criteria: prediction accuracy and computational efficiency. Prediction accuracy measures the number of outages detected before the permit went down (\emph{n\_early}). We also capture false negatives, where the system failed to detect a beam outage, and false positives, identified on validation data where no outage was expected, yet the system flagged one.

In real-world applications, accuracy alone isn’t enough; the model’s ability to integrate seamlessly with operations is equally critical. Therefore, in terms of computational efficiency, we compare the training time, inference time, and model size across different models.

\subsection{Beam Permit Prediction}
\emph{Prediction Accuracy:}
\Cref{tab:faultDetectionAccuracy} compares the prediction accuracy of all models on the test data, including $40$ operator-labeled outages, $40$ beam-permit-labeled outages, and $31$ non-outage instances. The results indicate that LSTM achieves the highest early detection rate among all models. The Transformer model ranks second in early detection, with a slightly lower false positive rate than LSTM. N-HiTS ranks third in early detection performance, offering a marginally lower mean squared error (MSE) compared to LSTM.  
Filtering out beam-permit fluctuations, instances where the beam permit drops for less than $10$ seconds, from the training data has proven effective in reducing the false positive rate and preventing the models from predicting random fluctuations in the beam-permit signal.

\begin{figure}[ht!]
    \centering
    \includegraphics[width=\columnwidth]{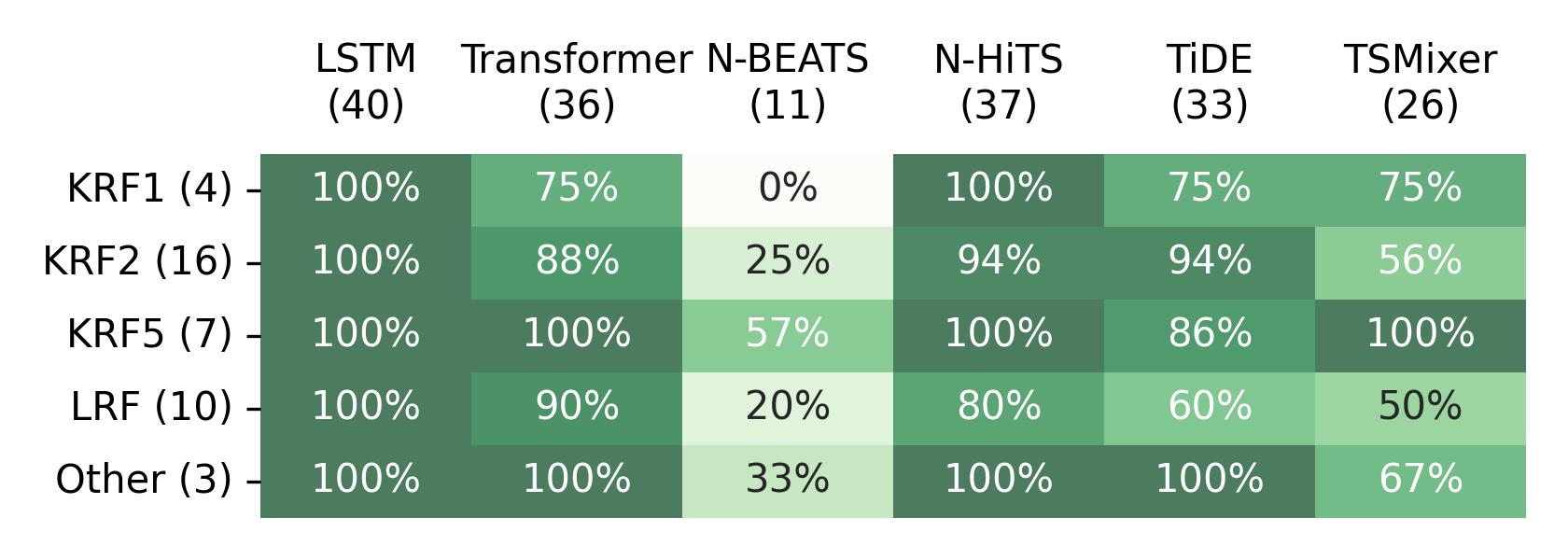}
    \caption{Model-wise detection rate of outage types.}
    \label{fig:det-rate}
\end{figure}
Figure~\ref{fig:det-rate} further characterizes model performance across different outage types on operator-labeled data. LSTM correctly predicted all instances of operator-labeled outages in the test dataset, while Transformer and N-HiTS architecture failed to early-predict some instances of \texttt{KRF1}, \texttt{KRF2}, and \texttt{LRF}. The ability to early predict an outage depends on two factors: (1) presence of a precursor signature (outages that evolve over time), and (2) the ability of a model to recognize those precursors. Abrupt outages (e.g. trips, manual intervention) are usually hard to predict because a precursor signature might not exist. 

However, since LSTM successfully predicted all operator-labeled outages, the missed early predictions by other models indicate their limited ability to detect precursors. For example, in 8 out of 30 instances of \texttt{KRF1}, \texttt{KRF2}, and \texttt{LRF}, either the Transformer or N-HiTS architecture was able to predict the fault early, but not both. This highlights the diversity in pattern recognition across models, which can be leveraged in future work by training an ensemble of architectures to combine their strengths and improve overall predictive performance.  

\noindent\emph{Computational Cost:}
Table~\ref{tab:performanceCharacterization} compares models’ performance in terms of training time, inference time, and model size—key factors for deployment considerations. From the inference perspective, Transformer is most efficient, whereas TFT performs the worst. Among the linear models, N-HiTS and N-BEATS are relatively computationally intensive, while TSMixer and TiDE are significantly lighter.

\input{compute_performance}





\subsection{Outage Labeling}
\begin{figure}
    \centering
    \includegraphics[width=0.98\columnwidth]{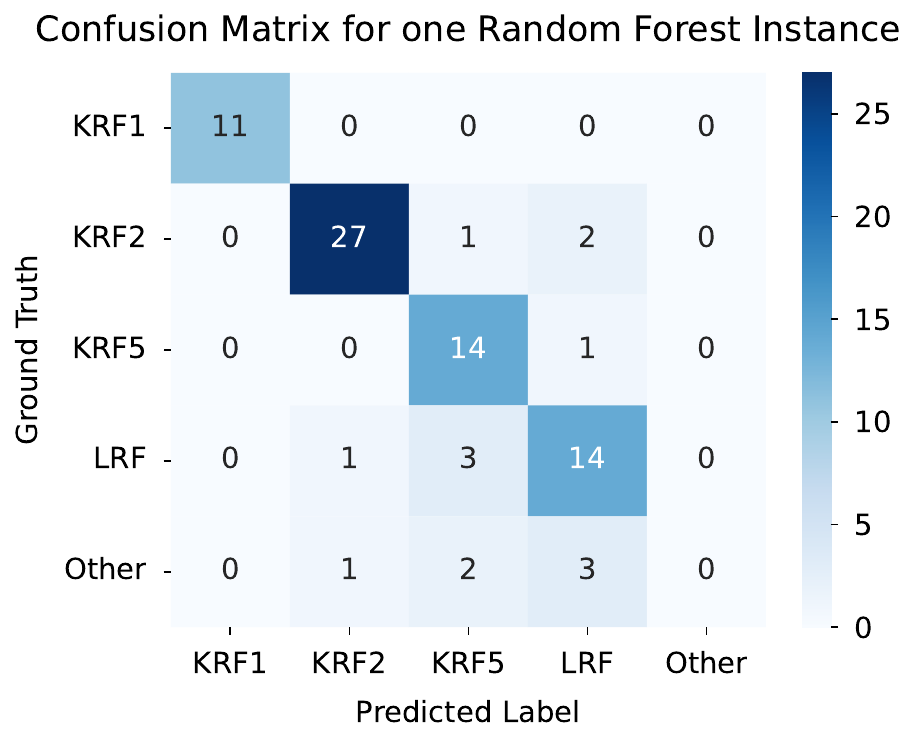}
    \caption{Confusion Matrix for Random Forest Classifier}
    \label{fig:rf-confusion}
\end{figure}

The random forest classifier was tested on the set of 80 operator-labeled outages. Due to the the relative smallness of the data, we use 8-fold cross validation when evaluating the classifier. Both the random shuffling for cross validation and the random state of the forest introduce variability to classification results. Over 100 iterations of cross-validation, our classifier had accuracy $\mathbf{0.821\pm0.021}$ and macro F1-score $\mathbf{0.691\pm0.018}$. A confusion matrix for one instance of cross validation is shown in \Cref{fig:rf-confusion}. 

\begin{figure}[ht!]
    \centering
    \includegraphics[width=\columnwidth, trim={0 0 0 0},clip]{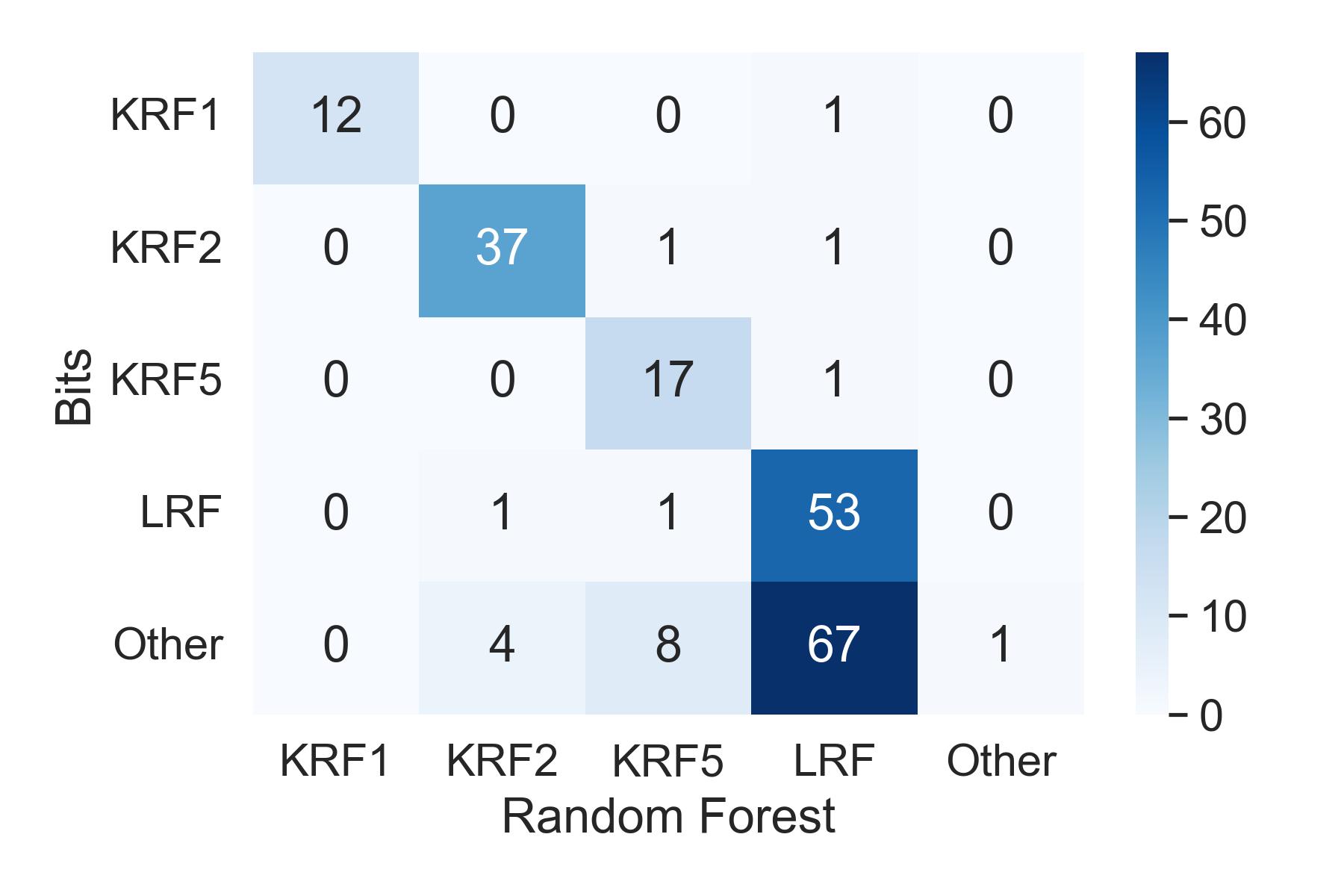}
    \caption{Comparison of RF Labeler with Bit Labeler}
    \label{fig:compare-confusion}
\end{figure}

Next, \Cref{fig:compare-confusion} shows a high degree of consistency between the pattern-based random forest and the summary information used in the bit-based labeler. Overall, these two methods reduce the number of unlabeled outages from 130 to only 1, and particularly very short outages can now be labeled. Nevertheless, we contend that more work is needed to reduce the number of off-diagonal elements in this matrix, outages on which the two methods disagree. The data suggests that we should be able to further refine the random forest-based labeling as more outages will be labeled by operators in the future.


\section{Discussion}
In this section, we present key insights from our analysis and the lessons learned, highlighting open problems that need to be addressed in future work to enable more efficient predictive operations in large facilities.
\subsubsection{LSTM outperforms SOTA DL architectures.}
Our analysis of SOTA DL architectures for beam outage prediction reveals that LSTM outperforms attention-based networks and linear networks across multiple dimensions.
Since these models are deployed for near-real-time inference, we considered the computation aspects as well. That said, transformers offer the added advantage of interpretability, and to capitalize on this, we plan to incorporate more efficient variants of transformer architectures in the future. 

\subsubsection{Random forest labeler demonstrated $\mathbf{82.1}$\% accuracy on operator-labeled outages.}
The random forest classifier was able to achieve a relatively high mean accuracy of $\mathbf{82.1}$\%. However, its Macro F1-score was less impressive, with a mean of $\mathbf{0.691}$. Here, we see that the random forest classifier has difficulty handling the high class imbalance of the ``Other'' class, which has only 3 observations.

However, an advantage of the random forest is fast training and inference time. The model took $0.394$s to complete training on all data and takes only $0.012$s to label one instance of data. The fast training time is beneficial for a deployed model that may need to be retrained or re-tuned in the future, and the fast inference time is necessary for real-time use. The model's memory usage is comparable to the networks used for inference, taking up 3.76 MB.
This model serves as an important cross-check of the bit-based labeler, with a high degree of consistency between the two. However, because of the easier interpretability of the random forest and direct connection to individual variables, we will build on this method in the future to study specific signatures of outages and further refine our utilization of pre-cursors for the outage prediction.  

\subsubsection{Interpretability is important.}
Operators are not only interested in predictions but also in the interpretability of those predictions. For example, which features at what specific times are correlated with the prediction. While significant progress has been made in interpretable ML for text and image data, gradient-based techniques like SHAP and Gradients struggle with multivariate time series (MTS). In our initial attempt, gradient-based saliency maps failed to distinguish important features, assigning high saliency to almost all features at each time step. This issue, also noted in recent literature~\cite{ismail2020benchmarking}, was partially addressed with Temporal Saliency Rescaling (TSR), though it is only designed for classification tasks. We look forward to future improvements in interpretability for MTS prediction.

\subsubsection{Significance of local normalization.}
In the current implementation, hourly data is normalized by centering each feature around the mean and scaling to unit variance. This ensures that input features are comparable. Since the normalization is applied locally to each file, it prevents the model from capturing changes in system settings over time, which helps in identifying abrupt local changes. However, this approach may obscure long-term shifts and trends. We plan to study this in future and explore alternative normalization techniques that can also preserve global trends and periodic patterns when required. This would enhance the model's ability to capture dynamics over extended time frames.

\subsubsection{Data loading is a time-consuming task.}
Time series data differs from other sequential data types like text, audio, and video in several ways. The order of data is crucial, limiting parallelism during loading and potentially causing load imbalance. Additionally, time series data involves overlapping windows, requiring more data to be loaded than needed, especially when data spans across multiple files. Lastly, unlike text or audio data, which have natural endpoints, time series data belongs to an infinite space, with patterns existing at various time scales. This results in high computational demands and creates a bottleneck in training and inference.

\section{Conclusion}
Operations at large complexes, such as FNAL, remain predominantly reactive even today. Advancements in AI, particularly in multivariate time series modeling, offer the potential to shift operators from being reactive to being \emph{predictive}. In this study, we studied SOTA multivariate time series prediction models to predict beam outages in near real-time. Our analysis on real-world data highlights the superior performance and efficiency of LSTM over SOTA linear and attention-based networks for this task. We also introduced a labeler that can automatically label the outages with 82.1\% accuracy. 
The beam prediction model and outage labeler are already deployed at FNAL control rooms for real-world impact assessment. 
In addition to addressing gaps related to interpretability, as identified in the Discussion section, the future work will explore advancements in the space of continual learning and transfer learning to achieve similar performance at scale and over time. 
\section{Acknowledgments}
This work was supported in part by the U.S. DOE Office of Science, Office of High Energy Physics, under award 76651: ``Machine learning for Accelerator Operations". Pacific Northwest National Laboratory is operated by Battelle Memorial Institute for the U.S. Department of Energy under Contract No. DE-AC05-76RL01830. It was also partly supported by Fermi Research Alliance, LLC under Contract No. DE-AC02-07CH11359 with the U.S. Department of Energy, Office of Science, Office of High Energy Physics. FERMILAB-CONF-24-0952-AD

\newpage
\bibliographystyle{aaai}
\bibliography{references}

\appendix
\section{Faults Description}

Table ~\ref{tab:faultsDescription} lists different operator labeled outages we encountered in this study. First column shows the location, and the second column gives the outage type information. One important observation we gathered from the list of operator supplied labels and our bit-label study is the discrepancies of operator labels. For instance, our analysis pointed out that “ZOV driver voltage”, “ZOV driver/voltage trip”, “ZOV Voltage Trip”, and “ZOV V” all correspond to the same outage type. Our bit labeler was able to recognize the same pattern in the bit device behavior and was able to cluster these types of errors into a single label.

\input{fault_list}

\begin{figure}[h!]
    \centering
    \includegraphics[width=0.6\linewidth]{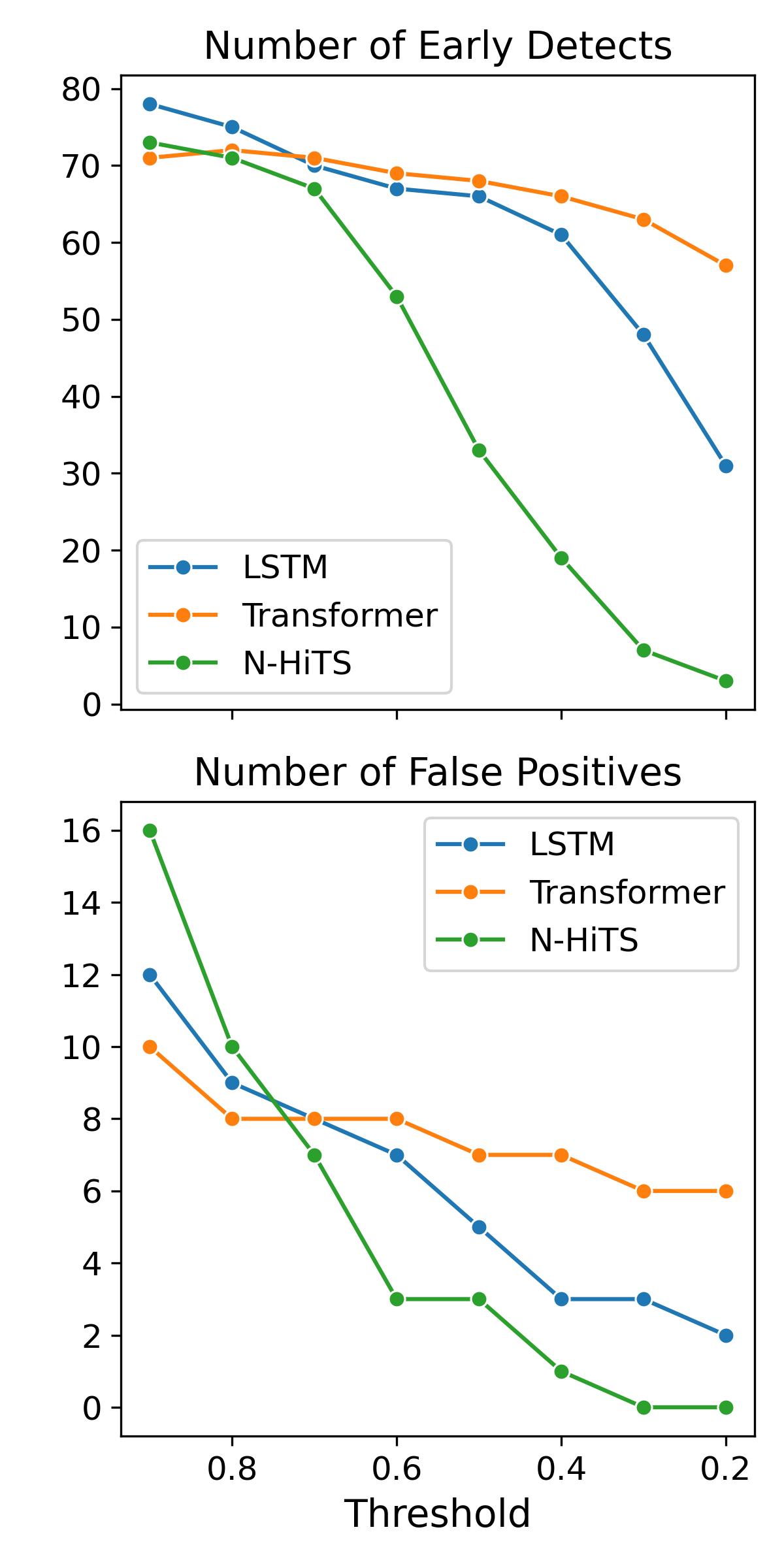}
    \caption{Threshold sensitivity.}
    \label{fig:threshold_sensitivity}
\end{figure}

\section{Sensitivity Analysis}
\label{app:sensitivity_analysis}
In this section, we discuss sensitivity of models trained for beam-permit prediction with respect to threshold values and data preprocessing parameters. 

\subsection{Threshold Sensitivity}
\Cref{fig:threshold_sensitivity} illustrates the sensitivity of the top three models (LSTM, Transformer, and N-HiTS) to varying threshold values, focusing on the number of outages detected early and the false positive rate. As the threshold decreases from 0.8, predictions that do not converge to zero are filtered out, leading to fewer early detections and a reduction in false positives. Conversely, increasing the threshold allows more predictions to pass, resulting in higher early detection rates and false positives.

Our analysis shows that Transformer and LSTM models exhibit greater robustness to threshold variations compared to N-HiTS. For both Transformer and LSTM, the early detection rate declines more steadily, indicating that their beam-permit predictions are closer to zero rather than experiencing sporadic fluctuations. In contrast, N-HiTS shows a sharper decline, suggesting that its predictions are less consistently close to zero and more susceptible to random dips.

\subsection{Preprocessing Sensitivity}
\Cref{tab:sensitivityAnalysis} presents the sensitivity analysis of all models with respect to three key factors: the size of the look-back window ($L_b \in \{45, 60, 90\}$), the gap between look-back and look-forward windows ($G \in \{0, 15, 60\}$), and the loss function ($Loss \in {BCEL, MAE}$). Here, $BCEL$ represents Binary Cross Entropy with Logits Loss, and $MAE$ corresponds to Mean Absolute Error. The \emph{Baseline} architecture refers to the original configuration with $L_b=30$, $G=30$, and $L_f=60$.
Following are some key takeaways.

\subsubsection{Impact of look-back window size:} 
Increasing the size of the look-back window allows models to utilize more historical data for beam-permit predictions. However, we observed a decline in the early detection rate for LSTM and Transformer models when increasing $L_b$ from $45$ to $90$. Conversely, for most linear models—except for TiDE—an improvement in early fault detection was generally observed. This highlights the significance of incorporating additional historical data to enhance the performance of linear models in predicting beam permits.

\subsubsection{Impact of gap:}
A gap between the look-back and look-forward windows enables the model to focus on future time steps. However, if the gap is too large, the model may fail to identify predictive patterns in the look-back data, leading to a reduction in its predictive power. Conversely, if the gap is too small, the model loses the advantage of forecasting and reduces the available time to take action. Therefore, finding the right balance is crucial. While the baseline architecture used a gap of $30$, we observed an improvement in the early detection rate for all models when the gap was increased to $60$, and a decline in performance when the gap was reduced. This suggests that, 2-4 seconds before an outage, the models are detecting disturbances in certain devices (\emph{precursors}) that are correlated with the upcoming outage. Future work will focus on identifying these precursors and providing operators with this information in advance. 

\subsubsection{Impact of changing the loss function:}
With the exception of N-BEATS, all other architectures demonstrate robustness to changes in the loss function. N-BEATS performs significantly better with BCEL and MSE. The high MSE value for BCEL indicates that the model is not optimized for MSE. In general, linear models tend to perform better with BCEL. Another notable observation is the increase in the false positive rate when using MAE.  


\input{sensitivity_an}

\end{document}

%% file: pred_acc.tex
\begin{table*}[!ht]
	\centering
	\caption{Prediction accuracy of models on operator-labeled beam outages.}
	\label{tab:faultDetectionAccuracy}
    \resizebox{0.98\textwidth}{!}{%
		\begin{tabular}{@{}lccccccccc@{}}
			\toprule
			& \textbf{MSE} & \textbf{n\_outages} & \textbf{n\_detected} & \textbf{Time diff.} & \textbf{n\_early} & \textbf{n\_late} & \textbf{False} & \textbf{n\_non} & \textbf{False}  \\
			& \textbf{(Test)} & & & \textbf{(in seconds)} & & & \textbf{Negatives (FN)} & \textbf{outages} & \textbf{Positives (FP)} \\
			\midrule
			\emph{Recurrent Networks}\\
            LSTM & 0.21 & 80 & 80 & -11.16 & 75 & 5 & 0 & 31 & 9 \\
			\emph{Attention Networks}\\
            Transformer & 0.21 & 80 & 79 & -9.62 & 72 & 7 & 1 & 31 & 8 \\
			\emph{Linear Networks}\\
            N-BEATS & 0.20 & 80 & 79 & -2.67 & 34 & 45 & 1 & 31 & 4 \\
			N-HiTS & 0.17 & 80 & 79 & -9.34 & 71 & 8 & 1 & 31 & 10 \\
			TiDE & 0.18 & 80 & 80 & -8.50 & 65 & 15 & 0 & 31 & 10 \\
            TSMixer & 0.19 & 80 & 76 & -5.40 & 49 & 27 & 4 & 31 & 8 \\
			\bottomrule
		\end{tabular}}
\end{table*}

%% file: compute_performance.tex
\begin{table}[t!]
	\centering
	\caption{Computational cost}
	\label{tab:performanceCharacterization}
	\resizebox{\columnwidth}{!}{%
		\begin{tabular}{lcccc}
			\toprule
			& \textbf{Model} & \textbf{\#Parameters} & \textbf{Train Time} & \textbf{Inference Time} \\
			& \textbf{Size (MB)} &  & \textbf{per epoch} & \textbf{per instance} \\
			&  &  & \textbf{(in mins)} & \textbf{(in seconds)} \\
			\midrule
			\emph{Recurrent}\\
            LSTM & 0.72 & 181K & 2.11 & 8.17 \\
			\emph{Attention}\\
            Transformer & 2.65 & 662K & 2.16 & 1.77 \\
			\emph{Linear}\\
            N-BEATS & 1720 & 496M & 4.87 & 33.38 \\
			N-HiTS & 542 & 135M & 2.57 & 17.39 \\
			TiDE & 2.57 & 642K & 2.46 & 3.17 \\
			TSMixer & 1.28 & 320K & 2.44 & 5.72 \\
			\bottomrule
		\end{tabular}
		}
\end{table}

%% file: fault_list.tex
\begin{table}[!ht]
	\centering
	\caption{List of operator-labeled outages in each category}
	\label{tab:faultsDescription}
    \resizebox{0.55\columnwidth}{!}{%
		\begin{tabular}{@{}lr@{}}
			\toprule
			& \textbf{Outage Label}\\
			\midrule
                \textbf{KRF1} & KRF1 CS Fault\\
			    
                \midrule
			\textbf{KRF2} & KRF2 CS Fault\\
			    
                \midrule
			\textbf{KRF5} & KRF5 CS Fault\\

                \midrule
			\textbf{LRF} & LRF1 FPGA Trip Sum \\
                & LRF1 trip \\
                & LRF2 Driver Anode OL \\
                & LRF2 reverse power \\
                & LRF3 FPGA trip \\
                & LRF3 FPGA trip sum \\
                & L3 O/I Trip \\
                & L3 Spark Trip \\
                & L3 ZOV Driver trip \\
                & L3 ZOV V \\
                & L3 ZOV Voltage Trip \\
                & L3 ZOV driver voltage \\
                & L3 ZOV driver/voltage trip \\
                & L4 High Voltage off \\
                & L4 VXI reboot\\
			    
                \midrule
			\textbf{Other} & KRF4 Gun Spark \\
                & KRF6 CS Fault \\ 
                & KRF6 reflected power fault \\
                & L:QPS312 issues \\
                & Roof leak on KRF7 PFN \\
			\bottomrule
		\end{tabular}}
\end{table}

%% file: sensitivity_an.tex
\onecolumn

\begin{longtable}{@{}lccccccccc@{}}
\caption{Sensitivity Analysis Results} \label{tab:sensitivityAnalysis} \\

\toprule
 & \textbf{MSE} & \textbf{n\_outages} & \textbf{n\_detected} & \textbf{Time diff.} & \textbf{n\_early} & \textbf{n\_late} & \textbf{False} & \textbf{n\_non} & \textbf{False} \\
 & \textbf{(Test)} & & & \textbf{(in seconds)} & & & \textbf{Negatives} & \textbf{outages} & \textbf{Positives} \\
\midrule
\endfirsthead

\multicolumn{10}{c}{\textit{Continued from previous page}} \\
\toprule
 & \textbf{MSE} & \textbf{n\_outages} & \textbf{n\_detected} & \textbf{Time diff.} & \textbf{n\_early} & \textbf{n\_late} & \textbf{False} & \textbf{n\_non} & \textbf{False} \\
 & \textbf{(Test)} & & & \textbf{(in seconds)} & & & \textbf{Negatives} & \textbf{outages} & \textbf{Positives} \\
\midrule
\endhead

\bottomrule
\multicolumn{10}{r}{\textit{Continued on next page}} \\
\endfoot

\bottomrule
\endlastfoot

\textbf{LSTM} & & & & & & & & & \\
\bfseries Baseline & \bfseries 0.21 & \bfseries 80 & \bfseries 80 & \bfseries -11.16 & \bfseries 75 & \bfseries 5 & \bfseries 0 & \bfseries 31 & \bfseries 9 \\
\cmidrule(lr){2-10}
$L_b=45$ & 0.21 & 80 & 80 & -9.07 & 70 & 10 & 0 & 31 & 6 \\
$L_b=60$ & 0.21 & 80 & 80 & -7.81 & 67 & 13 & 0 & 31 & 5 \\
$L_b=90$ & 0.24 & 80 & 74 & -6.84 & 63 & 11 & 6 & 31 & 5 \\
\cmidrule(lr){2-10}
$G=0$ & 0.15 & 80 & 80 & -1.60 & 25 & 55 & 0 & 31 & 2 \\
$G=15$ & 0.18 & 80 & 80 & -9.60 & 68 & 12 & 0 & 31 & 8 \\
$G=60$ & 0.25 & 80 & 80 & -12.31 & 78 & 2 & 0 & 31 & 10 \\
\cmidrule(lr){2-10}
BCEL$^{\dag}$ & 17.67 & 80 & 80 & -10.38 & 71 & 9 & 0 & 31 & 9 \\
MAE$^{\ast}$ & 0.22 & 80 & 80 & -10.05 & 72 & 8 & 0 & 31 & 12 \\

\midrule
\textbf{Transformer} & & & & & & & & & \\
\bfseries Baseline & \bfseries 0.21 & \bfseries 80 & \bfseries 79 & \bfseries -9.62 & \bfseries 72 & \bfseries 7 & \bfseries 1 & \bfseries 31 & \bfseries 8 \\
\cmidrule(lr){2-10}
$L_b=45$ & 0.20 & 80 & 79 & -4.74 & 48 & 31 & 1 & 31 & 4 \\
$L_b=60$ & 0.20 & 80 & 77 & -3.84 & 51 & 26 & 3 & 31 & 2 \\
$L_b=90$ & 0.23 & 80 & 76 & -3.39 & 49 & 27 & 4 & 31 & 2 \\
\cmidrule(lr){2-10}
$G=0$ & 0.16 & 80 & 79 & -3.61 & 43 & 36 & 1 & 31 & 4 \\
$G=15$ & 0.17 & 80 & 80 & -6.74 & 62 & 18 & 0 & 31 & 8 \\
$G=60$ & 0.21 & 80 & 80 & -10.27 & 77 & 3 & 0 & 31 & 6 \\
\cmidrule(lr){2-10}
BCEL$^{\dag}$ & 17.77 & 80 & 80 & -9.40 & 70 & 10 & 0 & 31 & 3 \\
MAE$^{\ast}$ & 0.24 & 80 & 80 & -10.80 & 75 & 5 & 0 & 31 & 7 \\

\midrule
\textbf{N-BEATS} & & & & & & & & & \\
\bfseries Baseline & \bfseries 0.20 & \bfseries 80 & \bfseries 79 & \bfseries -2.67 & \bfseries 34 & \bfseries 45 & \bfseries 1 & \bfseries 31 & \bfseries 4 \\
\cmidrule(lr){2-10}
$L_b=45$ & 0.22 & 80 & 80 & -6.13 & 60 & 20 & 0 & 31 & 4 \\
$L_b=60$ & 0.21 & 80 & 79 & -9.81 & 77 & 2 & 1 & 31 & 7 \\
$L_b=90$ & 0.24 & 80 & 75 & -7.28 & 67 & 8 & 5 & 31 & 5 \\
\cmidrule(lr){2-10}
$G=0$ & 0.16 & 80 & 77 & -0.46 & 25 & 52 & 3 & 31 & 2 \\
$G=15$ & 0.18 & 80 & 80 & -7.22 & 60 & 20 & 0 & 31 & 4 \\
$G=60$ & 0.23 & 80 & 79 & -11.34 & 76 & 3 & 1 & 31 & 8 \\
\cmidrule(lr){2-10}
BCEL$^{\dag}$ & 277.18 & 80 & 79 & -11.15 & 76 & 3 & 1 & 31 & 9 \\
MAE$^{\ast}$ & 0.23 & 80 & 80 & -9.02 & 74 & 6 & 0 & 31 & 11 \\

\midrule
\textbf{N-HiTS} & & & & & & & & & \\
\bfseries Baseline & \bfseries 0.17 & \bfseries 80 & \bfseries 79 & \bfseries -9.34 & \bfseries 71 & \bfseries 8 & \bfseries 1 & \bfseries 31 & \bfseries 10 \\
\cmidrule(lr){2-10}
$L_b=45$ & 0.19 & 80 & 78 & -7.16 & 66 & 12 & 2 & 31 & 6 \\
$L_b=60$ & 0.19 & 80 & 79 & -8.27 & 71 & 8 & 1 & 31 & 12 \\
$L_b=90$ & 0.20 & 80 & 75 & -8.86 & 74 & 1 & 5 & 31 & 20 \\
\cmidrule(lr){2-10}
$G=0$ & 0.12 & 80 & 79 & -9.21 & 72 & 7 & 1 & 31 & 19 \\
$G=15$ & 0.16 & 80 & 79 & -11.86 & 77 & 2 & 1 & 31 & 16 \\
$G=60$ & 0.20 & 80 & 80 & -11.53 & 79 & 1 & 0 & 31 & 14 \\
\cmidrule(lr){2-10}
BCEL$^{\dag}$ & 193.98 & 80 & 79 & -6.60 & 61 & 18 & 1 & 31 & 2 \\
MAE$^{\ast}$ & 0.18 & 80 & 80 & -11.43 & 77 & 3 & 0 & 31 & 15 \\

\midrule
\textbf{TiDE} & & & & & & & & & \\
\bfseries Baseline & \bfseries 0.18 & \bfseries 80 & \bfseries 80 & \bfseries -8.50 & \bfseries 65 & \bfseries 15 & \bfseries 0 & \bfseries 31 & \bfseries 10 \\
\cmidrule(lr){2-10}
$L_b=45$ & 0.19 & 80 & 79 & -8.65 & 69 & 10 & 1 & 31 & 11 \\
$L_b=60$ & 0.20 & 80 & 77 & -7.32 & 66 & 11 & 3 & 31 & 6 \\
$L_b=90$ & 0.23 & 80 & 76 & -4.93 & 58 & 18 & 4 & 31 & 4 \\
\cmidrule(lr){2-10}
$G=0$ & 0.14 & 80 & 79 & -5.86 & 56 & 23 & 1 & 31 & 7 \\
$G=15$ & 0.16 & 80 & 79 & -4.00 & 40 & 39 & 1 & 31 & 6 \\
$G=60$ & 0.21 & 80 & 77 & -11.62 & 75 & 2 & 3 & 31 & 8 \\
\cmidrule(lr){2-10}
BCEL$^{\dag}$ & 144.42 & 80 & 79 & -8.88 & 67 & 12 & 1 & 31 & 8 \\
MAE$^{\ast}$ & 0.20 & 80 & 78 & -5.96 & 53 & 25 & 2 & 31 & 7 \\

\midrule
\textbf{TSMixer} & & & & & & & & & \\
\bfseries Baseline & \bfseries 0.19 & \bfseries 80 & \bfseries 76 & \bfseries -5.40 & \bfseries 49 & \bfseries 27 & \bfseries 4 & \bfseries 31 & \bfseries 8 \\
\cmidrule(lr){2-10}
$L_b=45$ & 0.20 & 80 & 74 & -3.71 & 40 & 34 & 6 & 31 & 2 \\
$L_b=60$ & 0.20 & 80 & 79 & -6.07 & 61 & 18 & 1 & 31 & 8 \\
$L_b=90$ & 0.23 & 80 & 74 & -8.38 & 69 & 5 & 6 & 31 & 7 \\
\cmidrule(lr){2-10}
$G=0$ & 0.14 & 80 & 79 & -3.85 & 49 & 30 & 1 & 31 & 7 \\
$G=15$ & 0.16 & 80 & 79 & -4.27 & 50 & 29 & 1 & 31 & 5 \\
$G=60$ & 0.23 & 80 & 79 & -11.92 & 76 & 3 & 1 & 31 & 7 \\
\cmidrule(lr){2-10}
BCEL$^{\dag}$ & 45.04 & 80 & 78 & -4.50 & 48 & 30 & 2 & 31 & 4 \\
MAE$^{\ast}$ & 0.22 & 80 & 76 & -3.43 & 40 & 36 & 4 & 31 & 5 \\

\end{longtable}

\begin{tablenotes}
\item[$\dag$] BCEL: Binary Cross Entropy with Logits Loss (\texttt{torch.nn.BCEWithLogitsLoss})
\item[$\ast$] MAE: Mean Absolute Error (\texttt{torch.nn.L1Loss})
\end{tablenotes}

\twocolumn